\documentclass[conference]{IEEEtran}
\IEEEoverridecommandlockouts

\usepackage{cite}
\usepackage{amsmath,amssymb,amsfonts}
\usepackage{algorithmic}
\usepackage{graphicx}
\usepackage{textcomp}
\usepackage{xcolor}
\usepackage{subcaption}
\usepackage{booktabs}
\usepackage{pifont} 
\usepackage{ltablex}
\usepackage{colortbl}
\usepackage{epsfig}
\usepackage{url}

\newcommand{\cmark}{\textcolor{green!60!black}{\ding{51}}}
\newcommand{\xmark}{\textcolor{red}{\ding{55}}}

\def\BibTeX{{\rm B\kern-.05em{\sc i\kern-.025em b}\kern-.08em
    T\kern-.1667em\lower.7ex\hbox{E}\kern-.125emX}}
    
\begin{document}
%fix align name
\title{MS-DOTT: A Multi-Modal Dataset for Off-Road Traversability Estimation\\
\thanks{We acknowledge the support provided by Kotak IISc AI–ML Centre (KIAC) for the sensors used in this work.}
}

\author{
\begin{tabular}{ccc}

\begin{tabular}{c}
Shivani Subramanyan \\
\textit{Dept. of Aerospace Engineering} \\
\textit{Indian Institute of Science} \\
Bengaluru, India \\
shivanins@iisc.ac.in
\end{tabular}

&

\begin{tabular}{c}
Saksham Sharma \\
\textit{Dept. of Aerospace Engineering} \\
\textit{Indian Institute of Science} \\
Bengaluru, India \\
saksham.prob@gmail.com
\end{tabular}

&

\begin{tabular}{c}
Samahith S A\textsuperscript{*} \\
\textit{Dept. of Aerospace Engineering} \\
\textit{Indian Institute of Science} \\
Bengaluru, India \\
samahithsa@iisc.ac.in \\
\textsuperscript{*}Corresponding author
\end{tabular}

\vspace{1em}
\\[2.5em]

\multicolumn{3}{c}{
\begin{tabular}{cc}

\begin{tabular}{c}
Ahaan Kotian \\
\textit{Dept. of Aerospace Engineering} \\
\textit{Indian Institute of Science} \\
Bengaluru, India \\
ahaan.kotianstudent@gmail.com
\end{tabular}

\hspace{3em}

&

\begin{tabular}{c}
Suresh Sundaram \\
\textit{Dept. of Aerospace Engineering} \\
\textit{Indian Institute of Science} \\
Bengaluru, India \\
vssuresh@iisc.ac.in
\end{tabular}

\end{tabular}
}

\\[1.5em]

\multicolumn{3}{c}{

}

\end{tabular}
}

\maketitle

\begin{figure*}[t] 
\centering
\includegraphics[width=\textwidth]{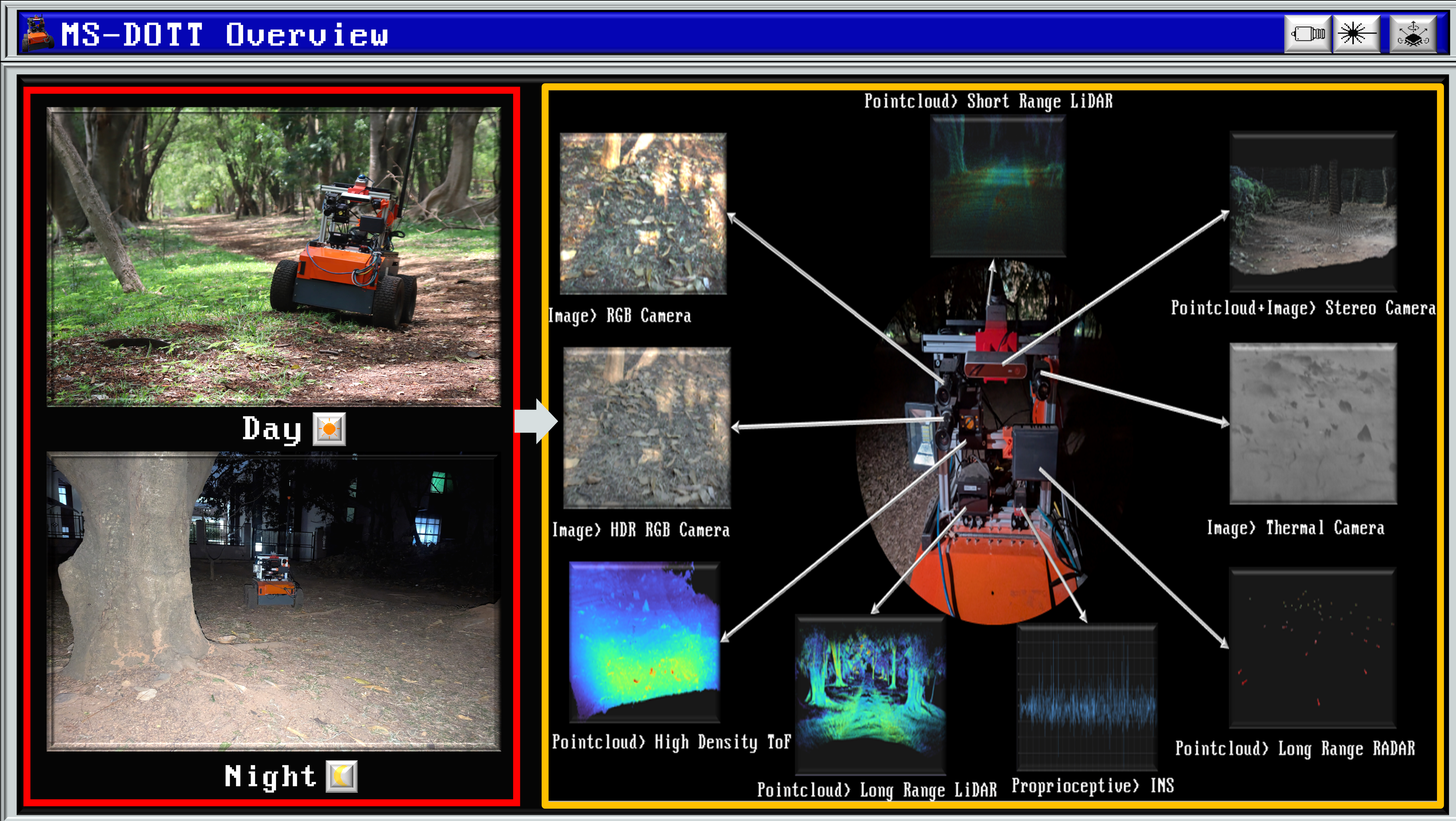}
\caption{MS-DOTT dataset overview. MS-DOTT consists of various complementary sensor data which allows for comprehensive terrain traversability estimation in both day and night conditions.}
\label{MS-DOTT}
\end{figure*}
% \vspace{10pt}

%gothru
\begin{abstract}
Autonomous robots are gaining traction in off-road environments, driven by their rising demand in agriculture, defense, and search-and-rescue operations. However, these environments are inherently unpredictable and navigation remains a formidable challenge given the complex environmental conditions including challenging terrain and varying illumination. Off-road autonomy requires robust perception across multiple spatial scales and imaging modalities, yet most sensing setups struggle to capture both far-field environmental context and near-field terrain detail simultaneously across various illumination conditions.  To facilitate  development of robust off-road perception and traversability estimation systems, we propose MS-DOTT (Multi-Sensor Dataset for Off-road Terrain Traversability) a comprehensive multi-modal off-road dataset with long-range 3D sensing and short-range, fine-grained perception to facilitate high-fidelity terrain understanding by combining proprioceptive signals for tracking vehicle dynamics with exteroceptive imaging and pointcloud data. The dataset includes trajectories recorded across nominal lighting, low-light, and no-light conditions with diverse 2D and 3D sensing for both long-range large-scale and short-range high-resolution sensing, as well as 4D RADAR measurements that provide velocity estimates for detected pointclouds. We demonstrate the utility of the dataset by adapting a SOTA multi-sensor fusion method for the purpose of traversability estimation. This dataset will aid the robotics community in advancing research and development of methods for traversability estimation, off-road perception and multi-sensor fusion in unstructured environments.
\end{abstract}


\section{Introduction}
Autonomous navigation has made significant progress in urban environments  \cite{ullrich2026toward, wang2025alpamayo, zhang2026deepsight, hu2025vision, huang2026automotunifiedvisionlanguageactionmodel, zhengdriveagent}, but reliable navigation in unstructured outdoor environments presents several unresolved challenges. Unlike on-road environments governed by lane markings, traffic rules and ambient lighting, off-road environments are highly unstructured and riddled with irregular topography, narrow trails and ambiguous boundaries. Obstacles and dense vegetation present in such environments introduce severe navigational hazards. In addition, variable surface conditions such as mud, gravel, and snow significantly affect vehicle mobility. The inherent variability in terrain and lighting makes navigation considerably challenging, limiting the direct applicability of conventional urban autonomous driving methods to off-road scenarios.
% The complete lack of structured infrastructure, combined with extreme variability in both geometry and lighting, greatly amplifies the complexity of perception and traversability estimation, precluding the direct transfer of urban autonomous driving methods to off-road scenarios.

Off-road autonomy has seen advances in perception and traversability estimation  \cite{min2026autonomous, sivaprakasam2025salon, 10160856, patel2024roadrunner, cai2025pietra, liang2025mosuautonomouslongrangerobot}, but translating these advances into robust, deployable real-world systems requires large-scale, multi-sensor datasets capturing both long-range perception and short-range terrain understanding. 
Existing datasets either focus on visual perception tasks such as 2D and 3D semantic segmentation, place recognition, and SLAM~\cite{8968283, 9561251,10611298, liu2024botanicgarden, knights2023wild}, or target specialised tasks such as self-supervised dynamics modeling~\cite{triest2022tartandrive} and off-road freespace detection~\cite{9812139}. 
However, several critical gaps remain. One, category-level semantics alone are insufficient to determine true traversability. For instance, regions identically labeled as "grass" or "vegetation" can vary drastically in actual traversability due to underlying differences in elevation, slope, and surface roughness. Second, datasets that attempt to capture 2D or 3D traversability ~\cite{park2026stonedatasetscalablemultimodal, sivaprakasam2024tartandrive, 9721297}, often rely on manually annotated labels that are costly to produce and scale or lack sufficient diversity in illumination and sensing modalities, resulting in limited coverage of challenging conditions.
Third, datasets with diverse illumination conditions~\cite{datar2025m2p2, liu2024botanicgarden, kim2025diter++} primarily focus on long-range perception for tasks such as SLAM, segmentation, end-to-end learning-based navigation and provide limited information about near-field terrain characteristics, which is needed for reliable local navigation. 
These limitations highlight the need for a multi-modal dataset that captures long-range perception and dense short-range terrain information to facilitate development of traversability estimation and off-road perception systems for safe off-road navigation.
To address this gap, we propose MS-DOTT (Multi Sensor Dataset for Off-road Terrain Traversability), a multi-modal off-road dataset that combines long-range 3D sensing with short-range, fine-grained perception across varying illumination conditions. The dataset integrates exteroceptive imaging and pointcloud data with proprioceptive signals for tracking vehicle dynamics, enabling long-range perception tasks while enabling high-fidelity understanding of near-field terrain. 
The dataset is collected using a rugged UGV platform, the Botsync Copernicus, equipped with an RGB camera, an HDR RGB camera, a thermal camera (LWIR), a high-density Time-of-Flight (ToF) sensor, a stereo camera, a short-range downward-facing 3D LiDAR, a long-range forward-facing 3D LiDAR, a forward-facing 4D RADAR, and a 9-axis IMU with multi-constellation GNSS.
To allow for simultaneous long-range perception and short-range terrain awareness to enable safety-critical operations, all the cameras and one 3D LiDAR are downward-facing, pointing to the near-field terrain in front of the robot.
The dataset includes trajectories recorded in both day and night conditions in a forest-like environment, with simultaneous recordings from nine unique sensors spanning ninety minutes and a total distance of 3000 m. Six trajectories were recorded during the day, while five trajectories were recorded at night.

A key objective of the dataset is to support the development and evaluation of traversability estimation methods. While semantic segmentation can provide useful traversability heuristics in relatively less-cluttered environments where trail boundaries are clearly visible, such heuristics can become unreliable in highly cluttered forest environments where the distinction between traversable and non-traversable terrain is ambiguous. 

The recorded trajectories capture such conditions, providing a basis for evaluating methods that can leverage richer multi-modal information for terrain traversability estimation. We validate the utility of this dataset by adapting a SOTA multi-modal traversability estimation method~\cite{sivaprakasam2025salon} using various combinations of sensors available in the dataset and discuss their respective performance in traversability estimation. To summarise, the primary contributions of MS-DOTT is as follows:

\begin{enumerate}
\item Dataset with simultaneous multisensor data for long-range perception and fine grained short-range perception, allowing for comparison across modalities.
\item Introduction of high density ToF and HDR-RGB sensing for fine grained perception in low light conditions.
\item Qualitative evaluation method for traversability estimation in off-road environments. 
\item ROS2 drivers for ToF sensing and GPU accelerated RGB pointcloud fusion. 
\end{enumerate}

\begin{table*}[t]
\centering
\caption{Comparison between Off-Road Datasets}
\label{tab:offroad_datasets}
\small % Sets a clean standard font size for full-width tables
\begin{tabular*}{\textwidth}{@{\extracolsep{\fill}} l c c c c c c c c c @{}}
\toprule
\textbf{Dataset} & \textbf{Thermal} & \textbf{RGB} & \textbf{ToF} & \textbf{LiDAR} & \textbf{Radar} & \textbf{IMU} & \textbf{Stereo} & \textbf{Platform} & \textbf{Illumination} \\ 
\midrule
RUGD~\cite{8968283} & \xmark & \cmark & \xmark & \cmark & \xmark & \cmark & \xmark & UGV & Day \\ 
RELLIS-3D~\cite{9561251} & \xmark & \cmark & \xmark & \cmark & \xmark & \cmark & \cmark & UGV & Day \\
CaT~\cite{9721297} & \xmark & \cmark & \xmark & \cmark & \xmark & \cmark & \xmark & ATV & Day \& Night \\ 
TartanDrive 2.0~\cite{sivaprakasam2024tartandrive} & \xmark & \cmark & \xmark & \cmark & \xmark & \cmark & \cmark & ATV & Day \\ 
ORFD~\cite{9812139} & \xmark & \cmark & \xmark & \cmark & \xmark & \cmark & \xmark & Car & Day \& Night \\ 
GOOSE~\cite{10611298} & \cmark & \cmark & \xmark & \cmark & \cmark & \cmark & \xmark & Car & Day \\ 
OORD~\cite{gadd2024oord} & \xmark & \cmark & \xmark & \xmark & \cmark & \cmark & \xmark & SUV & Day \& Night \\
M2P2~\cite{datar2025m2p2} & \cmark & \cmark & \xmark & \cmark & \xmark & \cmark & \cmark & UGV & Day \& Night \\
TOMD~\cite{11229344} & \xmark & \cmark & \xmark & \cmark & \xmark & \cmark & \cmark & UGV & Day \\ 
ORAD-3D~\cite{min2025advancingoffroadautonomousdriving} & \xmark & \cmark & \xmark & \cmark & \xmark & \xmark & \xmark & Car & Day \& Night \\ 
STONE~\cite{park2026stonedatasetscalablemultimodal} & \xmark & \cmark & \xmark & \cmark & \xmark & \cmark & \xmark & UGV & Day \& Night \\ 
BotanicGarden~\cite{liu2024botanicgarden} & \xmark & \cmark & \xmark & \cmark & \xmark & \cmark & \cmark & UGV & Day \\ 
DiTer++~\cite{kim2025diter++} & \cmark & \cmark & \xmark & \cmark & \xmark & \cmark & \xmark & Legged & Day \& Night \\
\midrule
\textbf{MS-DOTT} & \textbf{\cmark} & \textbf{\cmark} & \textbf{\cmark} & \textbf{\cmark} & \textbf{\cmark} & \textbf{\cmark} & \textbf{\cmark} & \textbf{UGV} & \textbf{Day \& Night} \\ 
\bottomrule
\end{tabular*}
\end{table*}

\section{Related Works}

\subsection{Off-road datasets}
    % Although existing off-road datasets have advanced field robotics, the majority predominantly feature data acquired using standard RGB cameras, IMUs, and LiDARs.
Existing off-road datasets have heavily driven progress in field robotics using standard sensors such as RGB cameras, IMUs, and LiDARs. RUGD~\cite{8968283} established the foundation by providing pixel-wise annotations for 2D semantic segmentation benchmarking, which RELLIS-3D~\cite{9561251} expanded into a multi-modal benchmark incorporating dense 3D pointcloud annotations. The CaT~\cite{9721297} dataset categorizes trails based on the mechanical overriding capabilities of different vehicle classes (e.g., sedans vs. off-road vehicles), arguing that standard semantic labels overlook vehicle-specific difficulty. 
For dynamics modeling, TartanDrive~\cite{triest2022tartandrive} utilized an ATV equipped with a stereo camera and shock sensors to train latent space models for efficient kinodynamic prediction, which TartanDrive 2.0~\cite{sivaprakasam2024tartandrive} scaled up with three additional LiDARs and self-supervised costmap generation via inverse reinforcement learning. ORFD~\cite{9812139} captured diverse terrains and illumination conditions using wide-angle cameras and LiDAR, introducing OFF-Net, a transformer-based architecture that fuses visual semantic features with 3D geometric surface information to detect freespaces. However, these datasets remain limited in terms of sensing modalities, illumination diversity, or short-range perception capabilities.

To address the above limitations, recent datasets have integrated diverse complementary sensor modalities to tackle off-road challenges. GOOSE~\cite{10611298} utilizes a LiDAR, RGB-NIR and thermal cameras, and six RADARs to provide granularly annotated pairs to benchmark advanced segmentation models. To handle extreme weather and lighting conditions, OORD~\cite{gadd2024oord} used a scanning RADAR to demonstrate robust place recognition and SLAM in the Scottish Highlands, evaluating relocalisation against GPS ground truth. Also targeting low-light conditions, M2P2~\cite{datar2025m2p2} leverages thermal, event, LiDAR, and stereo RGB cameras to prove that depth reconstruction is achievable in complete darkness.
TOMD~\cite{11229344} focuses on narrow trails under extreme illumination, employing a dynamic multiscale fusion model to integrate LiDAR and stereo camera features for path detection. ORAD-3D~\cite{min2025advancingoffroadautonomousdriving} combines LiDAR with five cameras in diverse terrains to establish baselines for freespace detection, 3D occupancy prediction, and VLM-based driving. BotanicGarden~\cite{liu2024botanicgarden} utilizes stereo cameras, LiDARs, and an IMU to demonstrate the importance of multi-sensor fusion for state estimation in natural environments, while STONE~\cite{park2026stonedatasetscalablemultimodal} leverages surround-view cameras, LiDAR, and RADAR through an annotation-free pipeline to generate geometry-aware 3D traversability maps. 
Despite the above advances, these datasets lack near-field fine grained sensing and simultaneous multi-sensor data for robust sensing in lighting constrained environments.  
MS-DOTT aims to overcome these limitations by introducing a versatile, UGV-based traversability dataset featuring a comprehensive sensor suite including thermal, RGB, HDR RGB, ToF, multi-range LiDARs, RADAR, stereo vision, and an IMU. Recorded across diverse off-road terrains in both day and night conditions, this dataset provides a comprehensive resource for developing resilient multi-sensor navigation systems capable of handling variable lighting and unstructured environments. A brief comparison summary is presented in Tab.\ref{tab:offroad_datasets}.

% \vspace{10pt}

\subsection{Traversability estimation}
Existing traversability methods range from semantic-based approaches like GANav~\cite{9810192}, which uses an attention mechanism for RGB terrain segmentation, to self-supervised frameworks like HDIF~\cite{10160856} that blend exteroceptive data with proprioceptive IMU feedback. More recent developments like Roadrunner~\cite{frey2024roadrunner} and its multi-modal extension Roadrunner M\&M~\cite{patel2024roadrunner} leverage self-supervised networks to predict safe speeds and traversability costs from visual-inertial cues. Similarly, Velociraptor~\cite{pmlr-v270-triest25a} utilizes visual foundation models and geometric mapping to build risk-aware costmaps without manual labels. SALON~\cite{sivaprakasam2025salon} advances these methods by enabling real-time online adaptation to out-of-distribution terrains through a bird's-eye-view (BEV) map fusing imagery and LiDAR. MS-DOTT utilizes SALON’s modular architecture to demonstrate the utility of the dataset in varied offroad conditions through fusion of various image and pointcloud data.  The performance of this fusion has been qualitatively demonstrated in Section~\ref{sec:result}.

\section{MS-DOTT}

% Robust autonomous navigation in off-road environments demands perception systems that
% can withstand a wide range of scene conditions, including variable lighting, adverse weather and ambiguous terrain. A single sensing modality cannot address all these challenges. Optical sensors are susceptible to glare and darkness, while geometric [CHANGE use a better term] sensors may fail on texture-less surfaces.
% By combining complementary sensor types the resulting multi-modal sensor suite achieves reliable coverage across scenarios where any individual sensor would fail. Figure \ref{MS-DOTT sensors} shows the sensor suite used for the MS-DOTT dataset.

\begin{figure}[h] 
\centering
\includegraphics[width=0.47\textwidth]{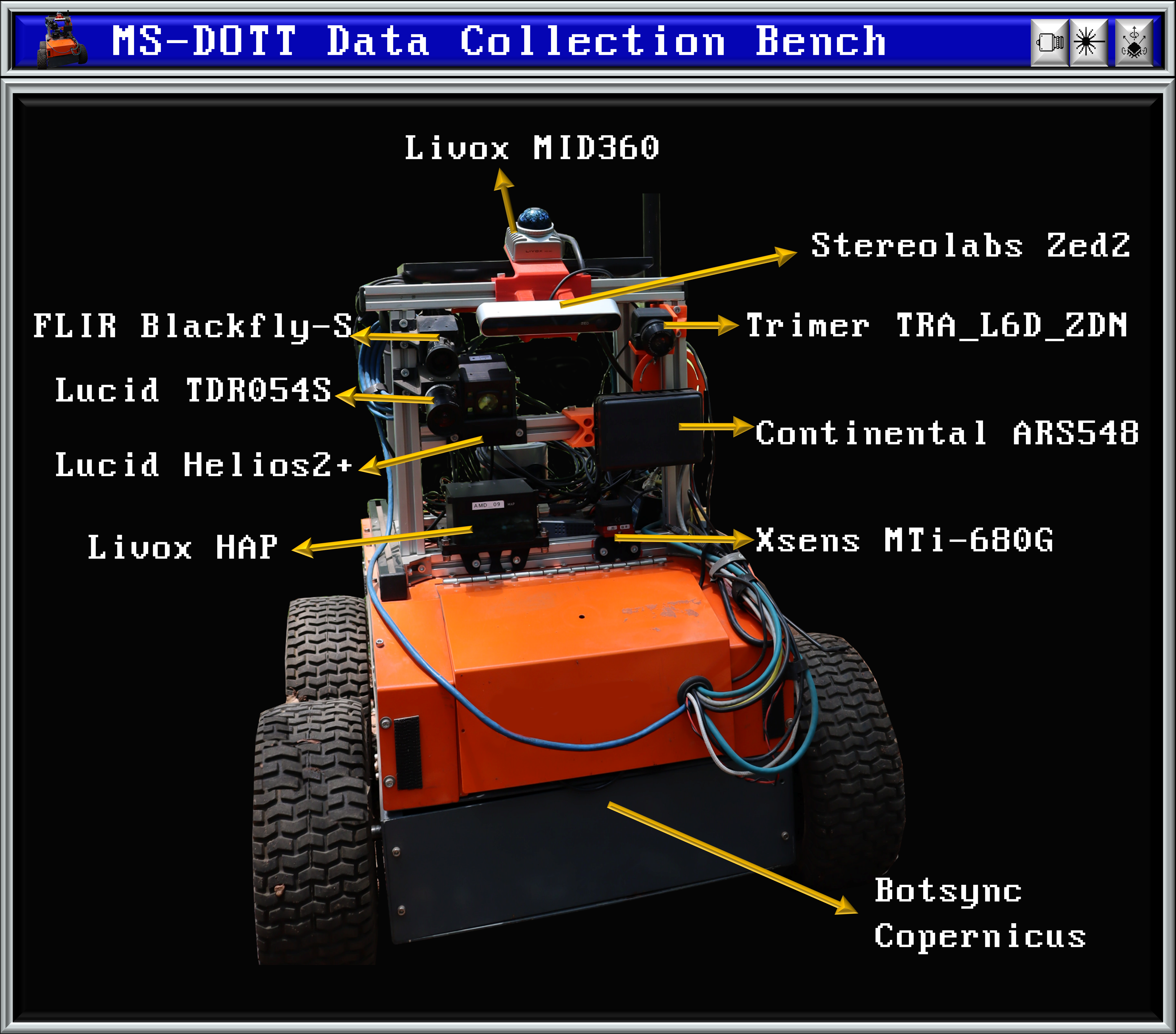}
\caption{MS-DOTT Data Collection Bench.}
\label{MS-DOTT sensors}
\end{figure}

\subsection{Data Collection Bench} 

The suite comprises a diverse array of sensors selected to enable complementary multi-modal sensing for long-range and short-range fine-grained perception as seen in Figure \ref{MS-DOTT sensors}.

\begin{table*}[t]
\centering
\caption{3D Sensor Specifications and Functional Roles within the MS-DOTT Dataset Platform}
\label{tab:lidar_specs}
\small
\setlength{\tabcolsep}{3pt} % Reduces horizontal padding so columns fit within textwidth
\begin{tabular*}{\textwidth}{|@{\extracolsep{\fill}}
    >{\centering\arraybackslash}m{0.16\textwidth}|
    >{\centering\arraybackslash}m{0.09\textwidth}|
    >{\centering\arraybackslash}m{0.09\textwidth}|
    >{\centering\arraybackslash}m{0.09\textwidth}|
    >{\centering\arraybackslash}m{0.11\textwidth}|
    >{\centering\arraybackslash}m{0.13\textwidth}|
    >{\centering\arraybackslash}m{0.25\textwidth}|}
\hline
\textbf{Sensor Name} & \textbf{Orientation} & Hz & \textbf{Range} & \textbf{FOV (HxV)} & \textbf{Mean points per scan} & \textbf{Purpose} \\ \hline

\textbf{Livox HAP} & Forward & 10 Hz & 150m & 120\textdegree $\times$ 25\textdegree & 44.91 $\times 10^3$ & Long-range environmental awareness, predictive path planning, and place recognition. \\ \hline

\textbf{Livox MID360} & Downward & 10 Hz & 40m & 360\textdegree $\times$ 52\textdegree & 19.83 $\times 10^3$ & Mid-range mapping, traditional terrain traversability baseline. \\ \hline

\textbf{ToF: Lucid Helios2+} & Downward & 15 Hz & 5m (8.3m max) & 67\textdegree $\times$ 51\textdegree & 233.16 $\times 10^3$ & Short-range, fine-grained terrain understanding and micro-obstacle detection. \\ \hline

\textbf{ARS548 RDI RADAR} & Forward & 10 Hz & 300m &  120\textdegree $\times$ 30\textdegree & 3 $\times 10^2$ & Weather agnostic perception \\ \hline

\end{tabular*}
\end{table*}

\subsubsection{LiDAR Sensors}
The platform uses two complementary LiDARs for 3D sensing. The forward-facing Livox HAP is a long-range, solid-state LiDAR used for long-range perception such as obstacle detection, path planning, and place recognition. Complementing it, the top-mounted Livox MID360 faces downward to provide terrain data specifically for traversability estimation. A summary of all LiDAR specifications is listed in Table: \ref{tab:lidar_specs}.
   
\subsubsection{Visual Sensors}
For rich visual and semantic perception under varied lighting conditions, the platform uses four downward-facing vision sensors. The FLIR Blackfly-S RGB camera provides standard RGB data for baseline terrain traversability estimation, while the ZED2 Stereo Camera provides both RGB imaging and stereo depth for short-range perception. To ensure operation in low-light scenarios, a Trimer TRA\_L6D\_ZDN thermal(LWIR) camera is integrated to capture critical terrain information through heat signatures. Also included is the Lucid Vision Labs TDR054S HDR RGB Camera featuring the IMX490 sensor which offers a high dynamic range of 123dB and additionally performs auto-tone mapping through Lucid Vision Labs's AltaView adaptive tone mapping\cite{altaview}. The aforementioned features allow the HDR camera to handle varying levels of illumination from sharp sunlight to low light scenarios effectively. A summary of all camera sensor specifications are listed in Table:\ref{tab:vision_specs}.

\subsubsection{IMU and RADAR}
The platform also integrates proprioceptive inertial sensors for vehicle state estimation and a long-range RADAR for weather agnostic perception. The forward-facing ARS548 RDI 77 GHz RADAR generates 4D pointcloud data with a range of up to 300 m and a field of view of 120° × 30°. It is capable of measuring point velocity via the Doppler effect, and it provides essential environmental awareness in challenging off-road scenarios as it is unaffected by adverse weather conditions. For high-accuracy localization, the platform incorporates a Xsens MTi-680G INS, which combines a 9-axis IMU with an L1/L2 RTK GNSS receiver to achieve a roll/pitch accuracy of 0.2° and a heading accuracy of 0.5°. IMU streams from Livox HAP and Livox MID360 are also recorded. Details of the IMU are specified in Table: \ref{tab:imu_specs}

\subsubsection{Helios ToF Sensor}
For short-range, ultra-dense terrain sensing, the platform integrates a Lucid Vision Labs Helios2+ ToF camera, which is facing downwards to capture small hazards like pebbles and roots directly in front of the vehicle as shown in Fig.~\ref{fig:lidar_benchmarks}. This sensor is rarely used in standard datasets. To integrate this sensor with the data collection bench, a custom ROS 2 driver has been developed to enable the robotics community to further utilize this sensor. Furthermore, a CUDA-accelerated algorithm is developed for real-time point painting, handling the projection of image data onto a 3D point cloud, which provides colored pointcloud for comparative analysis. Offloading point painting to the GPU reduces CPU computational load, allowing sufficient CPU resources to support multi-sensor data recording.

\begin{table*}[t]
\centering
\caption{Vision Sensor Specifications and Functional Roles within the MS-DOTT Dataset Platform}
\label{tab:vision_specs}
\small
\begin{tabular*}{\textwidth}{|@{\extracolsep{\fill}}
    >{\centering\arraybackslash}m{0.18\textwidth}|
    >{\centering\arraybackslash}m{0.09\textwidth}|
    >{\centering\arraybackslash}m{0.05\textwidth}|
    >{\centering\arraybackslash}m{0.09\textwidth}|
    >{\centering\arraybackslash}m{0.08\textwidth}|
    >{\centering\arraybackslash}m{0.32\textwidth}|}
\hline
\textbf{Sensor Name} & \textbf{Modality} &  \textbf{Hz} & \textbf{Resolution} & \textbf{FOV} & \textbf{Purpose} \\ \hline

\textbf{Trimer TRA\_L6D\_ZDN} & Thermal & 30 Hz & 640x480 & 30\textdegree $\times$ 23\textdegree & Short-range terrain understanding using thermal perception in low/no light. \\ \hline

\textbf{FLIR Blackfly-S} & RGB & 30 Hz & 1920x1080 & 48\textdegree $\times$ 36\textdegree & Short-range, baseline RGB data for terrain understanding in daytime. \\ \hline

\textbf{Lucid Vision Labs TDR054S} & HDR RGB & 10 Hz & 1920x1080 & 70\textdegree $\times$ 50\textdegree & Short-range, fine-grained terrain understanding robust to low-light and glares. \\ \hline

\textbf{ZED2 Stereo Camera} & RGB + Stereo depth & 10 Hz & 1280x960 & 110\textdegree $\times$ 70\textdegree & Maximum 3D detection range of 20m, auxiliary sensor for short-range terrain understanding. \\ \hline

\end{tabular*}
\end{table*}

\begin{table*}[t]
\centering
\caption{Proprioceptive Sensor Specifications and Functional Roles within the MS-DOTT Dataset Platform}
\label{tab:imu_specs}
\small
\setlength{\tabcolsep}{4pt}
\begin{tabular*}{\textwidth}{|@{\extracolsep{\fill}}
    >{\centering\arraybackslash}m{0.18\textwidth}|
    >{\centering\arraybackslash}m{0.09\textwidth}|
    >{\centering\arraybackslash}m{0.06\textwidth}|
    >{\centering\arraybackslash}m{0.28\textwidth}|
    >{\centering\arraybackslash}m{0.29\textwidth}|}
\hline
\textbf{Sensor Name} & \textbf{Modality} & \textbf{Hz} & \textbf{Specs} & \textbf{Purpose} \\ \hline

\textbf{Xsens MTi-680G} & INS & 400 Hz & 9-axis INS, L1/L2 RTK GNSS receiver, 0.2\textdegree\ roll/pitch accuracy, 0.5\textdegree\ heading accuracy  & Localization and vehicle inertial data \\ \hline

\end{tabular*}
\end{table*}

\begin{table*}[t]
\centering
\caption{Data Collection}
\label{tab:data_metrics}
\small
\setlength{\tabcolsep}{3pt}
\begin{tabular*}{\textwidth}{|@{\extracolsep{\fill}}
    >{\centering\arraybackslash}m{0.25\textwidth}|
    >{\centering\arraybackslash}m{0.12\textwidth}|
    >{\centering\arraybackslash}m{0.10\textwidth}|
    >{\centering\arraybackslash}m{0.08\textwidth}|
    >{\centering\arraybackslash}m{0.09\textwidth}|
    >{\centering\arraybackslash}m{0.13\textwidth}|
    >{\centering\arraybackslash}m{0.12\textwidth}|}
\hline
\textbf{Varying Terrain} & \textbf{Illumination} & \textbf{Trajectories} & \textbf{Distance} & \textbf{Runtime} & \textbf{Camera Frames} & \textbf{Pointclouds} \\ \hline

Muddy terrain, grassy fields, loose gravel paths, and undulating landscapes & Day and Night & 11 & ~3000m & 90 mins & 432,000 & 243,000 \\ \hline

\end{tabular*}
\end{table*}

\subsection{Calibration} 
Intrinsic camera calibration was performed using standard checkerboard patterns with the ROS2 package \cite{ros2camcalib}. For extrinsic alignment between all the sensors, transformation parameters were determined from their mechanical mounting locations on the sensor rig and were qualitatively evaluated to ensure sufficient alignment of sensors. The calibration data provided in the /tf\_static topic. Further the validity of the extrinsics is validated through demonstration of fusion in Section \ref{sec:result}.

\subsection{Data synchronization}
The dataset consists of multisensor data at varying frequencies. Due to the varying interfaces of the sensors and the relatively low speed of the vehicle, the sensors are software synchronized, we evaluate only the relative timestamping difference between the sensors. The median data proximity, i.e. the temporal difference between measurements collected by different sensors over $11$ trajectories is $8.71ms$ with a standard deviation of $63.44ms$. 

\subsection{Deployment}
The sensor suite is deployed on the Copernicus UGV, developed by Botsync. Copernicus is a rugged ground vehicle equipped with a differential-drive mechanism and designed for versatile deployments, making it suitable for data collection. The vehicle provides wheel odometry from its wheel encoder sensors. The complete configuration of the multi-modal sensor suite is illustrated in Fig.~\ref{MS-DOTT sensors}. The data acquisition platform is a Jetson Orin-AGX(64GB) running ROS2 Humble. The GiGE cameras are connected using a PCIe- Power-over-Ethernet expansion card. The LiDARs and RADAR are connected over Ethernet through a switch and the other sensors are connected through a USB hub, All sensor streams were recorded using a ROS2 composable node recorders, which reduces the inter process  conversion overhead of recording over DDS and all data was written directly onto a high-speed SSD (WD SN850X NVMe SSD). The data collection metrics are specified in Table \ref{tab:data_metrics}
        
% \subsection{Data Collection Areas and Metrics}
% The collection environments included muddy terrain, grassy fields, loose gravel paths, and undulating landscapes. Data was recorded at day to capture a broad range of natural lighting conditions, from direct overhead sunlight and at nighttime for low and no light conditions. 
% Collected across 11 trajectories, the dataset covers a distance of around 3,000 meters over 90 minutes of runtime. The resulting data comprises of 804,600 individual camera frames and 432,000 pointclouds.

\subsection{Post-processing}
Following raw data collection, several post-processing pipelines were executed to refine and enhance the dataset.
\subsubsection{Localization} For accurate vehicle tracking and environmental mapping, an inertial-LiDAR odometry framework (FAST-LIO) \cite{9697912} fuses the Livox HAP point clouds with Xsens IMU data.
\subsubsection{Pointcloud outlier removal} To clean the short-range spatial data, the raw point clouds from the Helios ToF sensor undergo Euclidean distance thresholding centered on the sensor origin, effectively removing out-of-range returns and sensor noise.
\subsubsection{Contrast enhancement} To maximize feature visibility in challenging environments, the downward-facing thermal camera data is processed using Contrast Limited Adaptive Histogram Equalization (CLAHE) \cite{mohammed2025contrast} for optimized contrast enhancement.

\subsection{Format of Dataset}
All sensor streams are stored in the MCAP format to ensure the compatibility with robotics perception stacks supporting ROS2 framework and additonally ZED2 data is stored in the native Stereolabs \texttt{.svo} format to allow for real time recording with low CPU overhead and low file size. Additionally Stereolabs provides tools to use \texttt{.svo} format with ROS2.

\begin{figure}[t] 
    \centering

    \begin{subfigure}[t]{\linewidth}
        \centering
        \includegraphics[width=\linewidth]{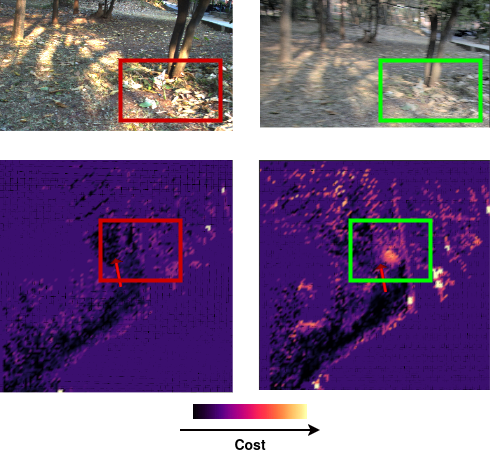}
        \caption{FLIR (Left) vs HDR (Right)}
        \label{fig:flir_hdr_glare}
    \end{subfigure}

    \vspace{0.4em}

    \begin{subfigure}[t]{\linewidth}
        \centering
        \includegraphics[width=\linewidth]{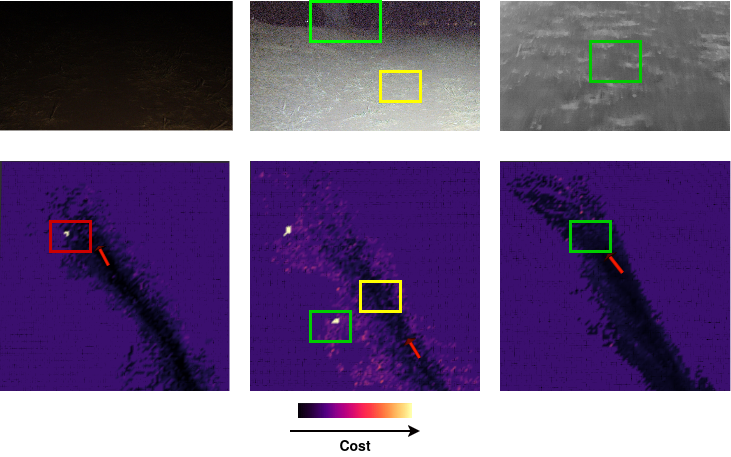}
        \caption{FLIR (Left) vs HDR (Middle) vs Thermal (Right)}
        \label{fig:flir_hdr_ir}
    \end{subfigure}

    \caption{\textbf{Performance comparison of different visual sensors across diverse lighting environments.} All costmaps (Bottom) are generated by fusion of Mid360 LiDAR and corresponding camera inputs (Up). (a) shows the effect of glares and shadows while (b) shows performance in nighttime conditions. Red arrow depicts pose of ego robot.}
    \label{fig:camera_benchmarks}
\end{figure}

\begin{figure}[t] 
    \centering
    \includegraphics[width=\linewidth]{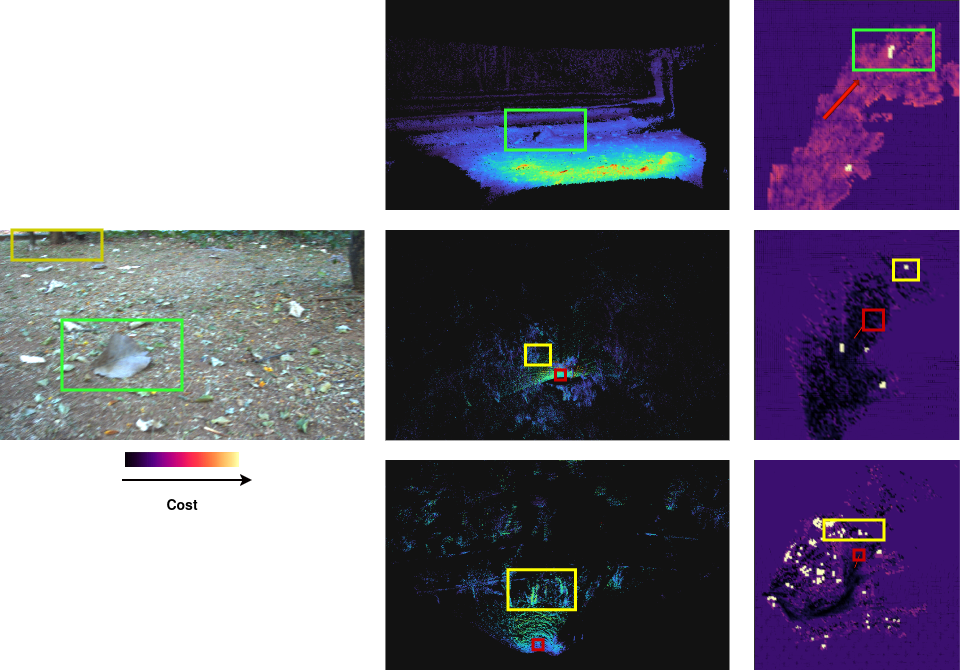}

    \caption{\textbf{Performance comparison of different LiDAR sensors for multi-scale traversability estimation.} The image on the left displays the reference FLIR camera frame. The images on the right illustrate the pointclouds and corresponding costmaps arranged vertically from top to bottom for the Helios, MID360, and HAP sensors, respectively. Red arrow depicts pose of ego robot.}
    \label{fig:lidar_benchmarks}
\end{figure}

\section{Experimental Results }
\label{sec:result}
We show the utility of MS-DOTT by discussing the performance of a SOTA traversability estimation method SALON~\cite{sivaprakasam2025salon} with different combinations of sensors present.
% , a framework designed for rapid online adaptation. SALON extracts pixel-level features from Visual Foundation Models (VFMs), compresses them via VLAD-based dimensionality reduction to preserve semantic properties, and projects them into a Bird's-Eye-View (BEV) map using LiDAR-camera calibration. Each BEV cell is then associated with exteroceptive terrain roughness—computed from the bandpower of the IMU—and robot speed, both of which are weighted to align with human intuition. For each cell, a predicted roughness distribution is estimated from the visual and inertial signals to provide a mean traversability cost. This allows SALON to refine its cost and speed maps within seconds of experiencing new terrain. Furthermore, a single user-provided label is sufficient to permanently register lethal obstacles, significantly reducing the need for extensive human teleoperation.
While SALON originally relies on DINOv2 \cite{oquab2023dinov2} for RGB inputs, standard vision transformers fail in low-light or nighttime conditions due to lack of features in the images. To extend SALON's traversability estimation to no light environments, we substitute DINOv2 with AnyThermal \cite{maheshwari2026anythermallearninguniversalrepresentations}, a thermal feature extraction backbone. AnyThermal is built upon the DINOv2 ViT-B/14 architecture and trained via knowledge distillation to align thermal features with frozen RGB visual priors. By swapping the vision backbone for AnyThermal, the remainder of the SALON pipeline proceeds unchanged, allowing for multisensor traversability costmap benchmarking.

\subsection{Camera benchmarks}
The qualitative results in Fig.~\ref{fig:flir_hdr_glare} show the impact of glare on costmap generation. Due to overexposure, the tree is omitted from the FLIR costmap (red) but is accurately captured in the HDR counterpart (green). This demonstrates the utility of HDR RGB camera to handle sharp and variable lighting while preserving information in the affected regions. Due to this, the costmaps generated are more expressive. Additionally, Fig.\ref{fig:flir_hdr_ir} qualitatively demonstrates performance of thermal and HDR RGB sensors against baseline RGB sensors (FLIR) in low-light conditions. The terrain is poorly identified and a clear area is falsely marked as high-cost (red) in the FLIR costmap while the tree is correctly identified as high-cost (green) and the traversable terrain in front of it is accurately labeled as low-cost (yellow) in HDR costmap. The clear terrain is also classified as low-cost (green) in IR costmap. This is because HDR and thermal handle low-light conditions robustly while FLIR yields highly underexposed images in low-light conditions.

\subsection{Multi-scale LiDAR benchmarks}

The qualitative results in Fig.~\ref{fig:lidar_benchmarks} highlight the utility of the ToF sensor in providing fine-grained, short-range terrain information. The ToF-based costmap correctly identifies near-field obstacles as high cost (green), whereas the other LiDAR-based costmaps (MID360 and HAP) fail to do so (red). The ToF sensor produces a more detailed and expressive local costmap proving its utility in short-range terrain understanding. Its limited range causes it to miss distant trees (yellow). Conversely, the longer-range LiDARs provide superior situational awareness and a broader navigation horizon. This difference in coverage yields small, medium, and large costmap footprints for the ToF, MID360, and HAP sensors, respectively. Ultimately, this unique combination of ToF and multi-range LiDAR data enables effective multi-scale traversability estimation.

%===============================================================================

\section{Conclusion}
\label{sec:conclusion}
This paper introduces MS-DOTT, a comprehensive multisensor dataset across varying off-road and illumination conditions. The dataset consists of eleven trajectories over a period of ninety minutes covering ~3000m of off-road terrain. The MS-DOTT dataset provides nine unique complementary sensor data, which allows for development and benchmarking of robust off-road autonomy. The primary purpose of the dataset is to aid with research on traversability estimation of off-road terrains (downward facing LiDAR,ToF and multiple cameras) and allow for benchmarking of localization through multisensor fusion(Forward facing LiDAR, Stereo camera and 4D RADAR). To this end we provide qualitative results on traversability estimation through multisensor fusion, demonstrating the utility of unique sensors used in the dataset, namely HDR RGB camera and high pointcloud density ToF sensor. The results clearly demonstrate the superior performance of HDR RGB camera over low light and sharp light conditions and demonstrates the capability of the ToF sensor to pick up fine grained details which are otherwise missed by traditional 3D sensing mechanisms. 
%===============================================================================

\section{Limitations}
Despite the comprehensive sensing capabilities of MS-DOTT, few key limitations must be acknowledged to guide future works. First, the exteroceptive sensors are strictly forward-facing and lack 360-degree surround-view coverage, in contrast to recent benchmarks like the STONE dataset\cite{park2026stonedatasetscalablemultimodal} which features multi-view surround cameras. Second, the data collection bench records multiple cameras, LiDARs, 4D RADAR, and INS, at varying frequencies which generates exceptionally large data volumes in a short amount of time.
This ultimately resulted in relatively short per trajectory length of $\sim250m$ at relatively low speed of $\sim1 m/s$, limiting the dataset's utility for modeling high-speed off-road dynamics or aggressive velocity variations over longer trajectories. 
Third, the ToF sensor is vulnerable to harsh sunlight hence, the pointcloud density is significantly reduced in the day sequences.
Finally, the dataset lacks hardware time synchronization between sensors and does not provide quantitative calibration re-projection errors.
Addressing these limitations through multi-view camera/LiDAR setup, expanded geographic testing, an improved storage pipeline, and hardware time synchronization remains a primary objective for future work. 

\section*{Acknowledgment}
We acknowledge the support provided by Kotak IISc AI–ML Centre (KIAC) for the sensors used in this work. 

% \bibliography{main.bib}
\bibliographystyle{IEEEtran}
\bibliography{main}
% \printbibliography

\end{document}